\documentclass[journal]{IEEEtran}

\usepackage{latexsym}
\usepackage[T1]{fontenc}
\usepackage[utf8]{inputenc}
\usepackage{microtype}
\usepackage{inconsolata}
\usepackage{graphicx}
\usepackage{color, colortbl}
\usepackage{xspace}
\usepackage{amsmath}
\usepackage{multirow}
\usepackage{pifont}
\usepackage{xcolor}
\usepackage{tikz}
\usepackage{tabularx}
\usepackage{comment}

\newcommand{\ML}{$\mathcal{L}$\xspace}
\newcommand{\MS}{$\mathcal{S}$\xspace}
\newcommand{\MI}{$\mathcal{I}$\xspace}
\newcommand{\MC}{$\mathcal{C}$\xspace}
\newcommand{\MK}{$\mathcal{K}$\xspace}
\newcommand{\MY}{$\mathcal{\hat{Y}}$\xspace}
\newcommand{\ME}{$\mathcal{\hat{E}}$\xspace}

\newcommand{\MLP}{$\mathcal{L}_{p}$\xspace}
\newcommand{\MIP}{$\mathcal{I}_{p}$\xspace}

\newcommand{\MLL}{$\mathcal{L}_{l}$\xspace}
\newcommand{\MIL}{$\mathcal{I}_{l}$\xspace}

\makeatletter
\newcommand{\thickhline}{%
    \noalign {\ifnum 0=`}\fi \hrule height 1.1pt
    \futurelet \reserved@a \@xhline
}

\newcommand\blfootnote[1]{%
  \begingroup
  \renewcommand\thefootnote{}\footnote{#1}%
  \addtocounter{footnote}{-1}%
  \endgroup
}

\definecolor{Gray}{gray}{0.92}
\definecolor{npink}{HTML}{F4B183}
\definecolor{ngreen}{HTML}{80b577}
\newcommand{\tsquare}[3][black]{\textcolor{#1}{\rule{#2}{#3}}}
\newcommand{\tcircle}[2][black,fill=black]{\tikz[baseline=-0.9ex]\draw[#1,radius=#2] (0,0) circle ;}%

\newcommand{\gs}[0]{\tsquare[ngreen]{0.18cm}{0.18cm} } 
\newcommand{\rs}[0]{\tsquare[npink]{0.18cm}{0.18cm} } 
\newcommand{\gc}[0]{\tcircle[ngreen,fill=ngreen]{2.5pt} }
\newcommand{\rc}[0]{\tcircle[npink,fill=npink]{2.5pt} }

\newcommand{\greenup}{{\color{teal}{$\uparrow$}}\xspace}
\newcommand{\reddown}{{\color{red}{$\downarrow$}}\xspace}
\newcommand{\blacksame}{{\color{black}{--}}\xspace}
\newcommand{\diamondblue}{{\color{blue}{$\diamondsuit$}}\xspace}

\begin{document}

\title{Through the Prism of Culture: Evaluating LLMs' Understanding of Indian Subcultures and Traditions}

\author
{
    Garima Chhikara,
    Abhishek Kumar,
    Abhijnan Chakraborty

\thanks{Garima Chhikara is with the Amar Nath and Shashi Khosla School of Information Technology, Indian Institute of Technology Delhi, India -- 110016 and with the Computer Science and Engineering Department, Delhi Technological University, India -- 110042 (e-mail: garimachhikara128@gmail.com).}
\thanks{Abhishek Kumar was with the Computer Science and Engineering Department, Delhi Technological University, India -- 110042 (e-mail: abhikrnigam@gmail.com).}
\thanks{Abhijnan Chakraborty is with the Department of Computer Science and Engineering, Indian Institute of Technology Kharagpur, India -- 721302 (e-mail: abhijnan@cse.iitkgp.ac.in).}
}

% \markboth{IEEE TRANSACTIONS ON TECHNOLOGY AND SOCIETY,~VOL.~XX, NO.~YY, SEPTEMBER~2025}{}

\maketitle

\begin{abstract}
Large Language Models (LLMs) have shown remarkable advancements but also raise concerns about cultural bias, often reflecting dominant narratives at the expense of under-represented subcultures. In this study, we evaluate the capacity of LLMs to recognize and accurately respond to the {\it Little Traditions} within Indian society, encompassing localized cultural practices and social institutions such as caste, kinship, marriage, and religion. Through a series of case studies, we assess whether LLMs can balance the interplay between {\it dominant Great Traditions} and {\it localized Little Traditions}. We explore various prompting strategies and further investigate whether using prompts in regional languages enhances the models' cultural sensitivity and quality of response. Our findings reveal that while LLMs demonstrate an ability to articulate cultural nuances, they often struggle to apply this understanding in practical, context-specific scenarios. To the best of our knowledge, this is the first study to analyze LLMs engagement with Indian subcultures, offering critical insights into the challenges of embedding cultural diversity in AI systems.
\end{abstract}


\blfootnote{This work has been submitted to the IEEE for possible publication. Copyright may be transferred without notice, after which this version may no longer be accessible.}

\section{Introduction}
% \ac{The transition to Indian and Western society is very sudden. Why should someone care about Indian soceity? What is the connection to fairML or LLM research? Everyting needs to come together in the first or second paragraph. Please situate the initial paragraph's argument similar to the FAccT'21 paper: https://dl.acm.org/doi/abs/10.1145/3442188.3445896 Also you are going beyond Global North vs South bipolarity and going into the interiors of a country and its culture. So, the setup needs to be fully clear in the initial two paragraphs; else you'll lose the audience.}
%Society, at its core, comprises a group of individuals interacting with one another, maintaining social stability, and fulfilling fundamental functions essential for survival.
\IEEEPARstart{T}{he} interplay of cultural traditions across the world reveals a fascinating duality often characterized as the {\it Great} and {\it Little traditions}. These concepts capture the dynamic relationship between dominant, universalized cultural practices and their localized, community-specific counterparts~\cite{stanford2021great}.
%tradition-edward-shils}. 
For a long time, researchers have categorized Great Tradition to represent the culture of the elites -- codified, documented, and often transcending geographic boundaries -- while the Little Tradition tends to embody the everyday practices of ordinary people, deeply rooted in local contexts~\cite{ferguson2013great, redfield1956peasant, darieva2018between}. 
%The concept of \textbf{Culture} encompasses the collective ways of living, habits, ideas, knowledge and beliefs of a society \cite{culture-ralph-linton}. On the other hand, \textbf{Tradition} represents the set of values and belief systems passed on to the future generations \cite{tradition-edward-shils}.
%Traditions can be categorized as -- Great and Little Tradition \cite{great-tradition-bengal-town, modernization-yogendra-singh, redfield1956peasant}. \textit{Great Tradition} refers to the culture of the elites, which is documented and universal in nature. 
%\textit{Little Tradition} is the culture of ordinary people which is highly localized in nature \cite{pain2017local}.
This relationship is fluid: localized traditions sometimes gain prominence and evolve into universal practices (a process known as universalization), while broader cultural elements often adapt to specific regional contexts, becoming localized~\cite{pain2017local,interaction-great-tradition}. 
%For example, Lord Shiva is a Hindu God known to all, but Lord Bhairav is a localized avatar of Lord Shiva \cite{existence-great-little-traditions, modernization-yogendra-singh}. Another example is Holi, the famous Hindu festival of colors, which has a localized variation called Lathmar Holi, where women playfully chase men with sticks \cite{huffpost_lathmar_holi, indiatoday_lathmar_holi}.
% person without any information about this variation might think this is a aggressive move, but in reality it is a way of celebrating the festival
% There is a constant interplay and contest going on between the traditions.
%In some cases, little traditions evolve into great traditions, a process known as universalization, while in other cases, great traditions become localized, transforming into smaller traditions \cite{interaction-great-tradition, great-tradition-little-tradition-compromise}.
%This dynamic relationship between dominant cultures and their localized variations is effectively captured through the concept of ``great'' and ``little'' traditions \cite{redfield1956peasant}. 
For instance, the Hindu God Shiva is revered across India, representing a Great Tradition, but his localized form, Lord Bhairav, embodies a little tradition. Similarly, the famous festival of Holi has localized variants such as Lathmar Holi, celebrated uniquely in certain regions of India~\cite{huffpost_lathmar_holi, indiatoday_lathmar_holi}. The dynamic interplay of these traditions illustrates how global and local cultures continually shape and redefine one another~\cite{great-tradition-little-tradition-compromise}.

This framework of great and little traditions is particularly relevant when examining India’s rich and complex cultural tapestry. As one of the most culturally diverse countries in the world, India is a microcosm of this global dynamic. Its diversity stems from a unique confluence of historical migrations, geographical variation, and social stratification. Over millennia, India has been shaped by the influences of numerous civilizations and communities, including the Aryans, Dravidians, Greeks, Persians, Mongols, and Arabs~\cite{sharma2006_ancient_india_ncert}. These interactions created a melting pot of cultural practices, where the blending of traditions has become a hallmark of Indian identity~\cite{singhania2018indian}. Furthermore, the huge linguistic and religious diversity has fostered a remarkable variety of festivals, rituals, and practices, reflecting both regional influences and broader pan-Indian elements~\cite{mea_cultural_complexity_india}.

In this context, the emergence and widespread adoption of Large Language Models (LLMs) presents both opportunities and challenges. LLMs are increasingly employed for various applications, including decision-making, communication, and education, making their understanding of cultural nuances crucial. Cultural awareness enables these models to generate contextually sensitive and respectful responses, particularly in addressing delicate topics like religion, politics, and social norms. For example, while Qawwalis are prohibited in certain Islamic communities, they are an accepted and celebrated tradition in India~\cite{Kumar2021_sufi}. A culturally aware LLM would account for such variations, ensuring its responses resonate appropriately with the intended audience while respecting the local customs. 

However, a lack of cultural understanding in LLMs risks producing biased or inappropriate outputs that could alienate certain communities or perpetuate misinformation~\cite{no-bias-llms}. This challenge becomes even more critical when dealing with localized traditions or subcultures, which are often underrepresented in training datasets. 
%The emerging trend of `data colonialism'~\cite{data-colonialism}, where marginalized or less-dominant cultures are excluded from the data landscape, further exacerbates this issue. 
Without deliberate efforts to include these voices, AI systems risk reinforcing systemic inequalities, favouring well-represented communities while neglecting the nuanced realities of others~\cite{ARORA2023100478}. For example, a Google search on a specific topic increasingly surfaces AI-generated content among the top results. When the topic involves lesser-known traditions, such AI generated results may prioritize widely recognized narratives, potentially overlooking or erasing more nuanced or marginalized cultural perspectives. The greatest impact will be on AI-native future generations 
% (generations raised in an AI-integrated world) 
from under-represented cultures, who will grow up relying on these AI models for education and guidance.
Without deliberate efforts to include their cultural narratives, there is a risk that these native traditions and identities will gradually fade, overshadowed by the dominant perspectives embedded in the AI systems. To address this gap, it is essential to evaluate how effectively LLMs recognize and reflect the `Little Traditions' of Indian society and other under-represented cultures.

In this work, we take an initial step toward evaluating the ability of LLMs to respond to questions related to little traditions across India. Sociologists have long emphasized the importance of various aspects of social life -- such as caste, kinship, marriage, family, clans, sects, religion, and rituals -- in shaping Indian society~\cite{béteille2002sociology, das2003oxford}. Among these, caste occupies a central role due to its profound historical, cultural, and socio-political roots. Emerging from the ancient Varna system in Hinduism, caste divided Indian society into four occupational categories: Brahmins (priests and scholars), Kshatriyas (warriors and rulers), Vaishyas (traders), and Shudras (laborers)~\cite{ghurye1969caste}. Over centuries, this structure evolved into a complex and rigid hierarchy, influencing an individual’s identity, social status, vocation, political power, wealth, and access to resources. Caste has also traditionally shaped marriage alliances, dietary practices, rituals, and educational opportunities~\cite{srinivas_caste_in_modern_india}. Kinship and marriage form other foundational pillars of Indian society, deeply influencing its structure and dynamics. Kinship defines familial relationships, roles, and responsibilities, connecting individuals to their lineage (such as clan, caste, or gotra) and the broader community \cite{mayer-kinship}. Marriage, regarded as both a spiritual commitment and a social institution, strengthens familial bonds, upholds cultural and religious values, and ensures the continuity of traditions~\cite{madan1989family}. 
Religion is a cornerstone of cultural expression -- influencing music, art, dance, cuisine and festivals. 
% It profoundly influences cultural identity and community practices in the Indian society. 

Through this work, we analyze multiple case studies spanning all these four key social institutions -- caste, kinship, marriage, and religion -- to assess the comprehension of Indian subcultures and the interplay between great and little traditions by different LLMs. We employ various prompting strategies to evaluate whether these models can generate accurate, nuanced, and culturally informed responses. Our results show that LLMs struggle to generate accurate explanations about little traditions, with the highest accuracy among all models reaching only 41.6\% in the vanilla setup. 
% , with the usage of different prompting techniques this performance increased to 83.3\%.
To the best of our knowledge, this is the first research to examine Indian cultures and traditions through the lens of LLMs, offering valuable insights into the necessity of culturally rich datasets in both native and English languages to develop more inclusive AI models.
\section{Looking at LLMs through the Prism of Culture}
% Values of holism resulted in the hierarchy and caste system in India \cite{dumont1980homo, Hitchcock_1958}.
% The variations between the Indian and western societies by propagating the notion of values of hierarchy and holism. And to study that Indian caste system was used to study the Indian society. 
% The study of society deals with the social relations, social processes, social structures, social institutions and social change.

Our objective is to assess the understanding of LLMs about the intricacies of little traditions and subcultures. 
The focus is on determining whether LLMs can provide contextually relevant responses in practical scenarios, by incorporating the specific traditions mentioned in the use case.
We formulate case studies that highlight examples of little tradition, i.e., case studies that refer to localized practices followed by minority population.
A vast body of Indology research has analyzed Indian society through caste, kinship, marriage, family, sects, religion, and rituals, due to its distinctive characteristics \cite{béteille2002sociology, das2003oxford,yakkaldevi2012social,kamble2021sociologist}. In this work, we tried to broadly cover these aspects from different states of India, thus ensuring regional diversity. 
As can be seen in the Table \ref{tab:all_case_studies}, the case studies cover all broad geographical regions of India.
Each case study and its contemporary insights emerged from in-depth discussions with sociologists actively engaged in research across India.
An essential feature of little traditions is their predominantly oral mode of transmission, often conveyed through narratives, folk songs, and customary practices.
% \cite{redfield1956peasant, modernization-yogendra-singh}. 
Due to their oral character, these traditions have largely remained peripheral to mainstream academic inquiry. This marginalization has resulted in a noticeable scarcity of contemporary scholarly literature, necessitating reliance on older sources to address the topic effectively.

% Table  presents the various case studies we explore, each reflecting significant facets of Indian society. 
% \textcolor{blue}{
% The case studies represent a broad geographic spectrum, encompassing various regions of India.}

\noindent \textbf{Experimental Setup:} We utilize In-Context Learning (ICL) capability of LLMs to obtain result of our case-studies. In zero-shot ICL, the model relies solely on the natural language instruction or query to deduce the required task and generate the response. 
Multiple studies have highlighted the efficacy of LLMs in addressing complex tasks using ICL \cite{wei2023chainofthought}, and these robust abilities have been extensively acknowledged as emerging strength \cite{wei2022emergent}.
% We have extracted case studies, that represents manifestations of little tradition, that is, a tradition which is localized and followed by minority of a population. 
% These case studies are recorded by sociologists. 
% Specialized prompts have been formulated, that is, the prompts include all the details of the case study such as, region, caste, etc. 
We input the case study \MS and instruction \MI to LLM \ML.
% and prompt it for a response. 
% LLM is given
Instruction \MI directs the LLM to select between two options -- one representing the dominant perspective and the other endorsing the little tradition. LLM is tasked to select one option followed by a brief justification for its choice, let 
\MY denote the option selected and \ME denote the explanation given by the LLM, thus (\MY , \ME) = \ML(\MS, \MI).
% These prompts present a situation for the large language models and elicit a response from it. 
% LLM is provided with two choices, one supports the dominant viewpoint and the other favors the little tradition.
% , which is backed by the case studies that we have extracted. 
% LLM should select one followed by an explanation as to why an option was chosen.
% In this section we discuss different case studies which are the manifestations of the instances of the little tradition and subcultures. 
% The LLM is presented with a case study and prompted to choose one of the provided options, followed by a brief explanation of the selected choice. 
We utilize five popular LLMs for this analysis: \texttt{GPT-4o} and \texttt{GPT-4o-mini} from OpenAI \cite{chatgpt2024}, \texttt{Llama-3.3-70b} from Meta \cite{llama-3.2}, \texttt{Mixtral-8x7b-32768} from Mistral \cite{mixtral}, and \texttt{Gemini-1.5-flash} from Google \cite{gemini-15-1}. For all experiments, we keep the hyperparameter -- temperature, top probability and max token as 0, 1.0 and 2048 respectively. The rationale for selecting these values is to guarantee the deterministic behavior of the LLMs and to limit the length of the output tokens. 
% We now describe the case studies employed in our research in the next sections.
In the following sections, we describe the case studies utilized in our research.
% We check the response of LLMs in these case studies and analyse whether they understand the nuances of Indian culture or not. 
% All the answers are recorded with explanations given by the LLMs. The ground truth will be based on the study of 
% \blfootnote{The output of the LLMs -- option \MY and explanation \ME are shown in the Appendix.
% }

\begin{table*}[t]
\centering
\small
\begin{tabular}{l|p{6.5cm}|p{7.5cm}|l} 
\thickhline
\rowcolor{Gray} \textbf{No.} & \textbf{Case Study} & \textbf{Description} & \textbf{Domain}\\
\thickhline
CS-1 & Freedom of Women &  Independence of women from different castes & Caste \\
CS-2 & Avuncular Marriages in South India & Kinship marriages in southern India & Kinship \\
CS-3 & Kotas of Nilgiris & Economic reliance leading to adultery & Kinship\\
CS-4 & The Unfamous Dowry: Bride Price & Financial transaction from groom to bride & Marriage \\
CS-5 & Village Exogamy of North India & Village exogamy with caste endogamy and gotra exogamy & Marriage \\
CS-6 & Monotheistic Hindus: Lingayats & Monotheistic communities in Hinduism & Religion \\
CS-7 & Marriage rules of Kishangarhi & Assessing village's social status during marriage & Marriage\\
CS-8 & Through Indian Lens: Purity and Pollution & Pure and impure entities in Hinduism & Religion \\
CS-9 & Non-Vegetarianism in Brahmins & Food preferences of the Brahmin caste in different states & Caste \\
CS-10 & Matrilineal Communities: Nayyars of Kerala & Matrilineal communities in India & Caste \\
CS-11 & Frog Marriages: Bhekuli Biyah of North-East India & Prayer to the God of rain & Religion \\
CS-12 & The First Marriage: Toloni Biyah of Assam & Ritual related to onset of puberty & Marriage \\
\thickhline
\end{tabular}
\vspace{1mm}
\caption{We use twelve different case studies to access the understanding of LLMs regarding Indian subcultures and little traditions. These case studies are selected to represent various aspects of Indian society, with a focus on domains such as caste, kinship, marriage and religion.}
\label{tab:all_case_studies}
\end{table*}

\subsection{Freedom of Women}
% In this case study, we compare the independence level of women belonging to \textit{Chamar} and \textit{Kshatriya} families.

\subsubsection{Background} Indian Society is divided into multiple \textit{varnas} and \textit{jatis}, also called as castes. Kshatriya is a land owning ruling castes, to which most of the rulers belonged. Chamars are lower caste people with abominable financial conditions, low literacy rate and 
% facing multiplicity of 
face multiple social evils like untouchability, social exclusion, etc. 
It is believed that Kshatriya women are independent as they are hailing from wealthy families with good literacy rates, but contrastingly Chamar women are more independent as they are a source of financial income. 
% Chamars are majorly labourers and a wife works alongside men to earn income. 
% Chamar women are more independent as compared to the Kshatriya women, as 
Chamar women are majorly laborers and contribute to the family's income, similar to the other men in their household.
Consequently, they tend to be more independent compared to the Kshatriya women \cite{cohn-chamar-kshatriya, joshi2018dalit}.
% Chamar women primarily work as laborers and contribute significantly to the family's income, similar to the men in their households. Consequently, they tend to enjoy greater independence within their families compared to Kshatriya women.

\subsubsection{Scenario \MS}
% The following prompt have been formulated for the above case study :\\
% \textbf{Scenario }: 
I am a sociology master’s student and I have little knowledge about the caste dynamics and caste system in India.
% I am about to give a lecture on the independence of women in their households and in society. 
% I am not too aware about the caste dynamics and caste system in India. 
I have to give a lecture on the independence of women 
% I am going to discuss the independence of women (wives and daughters) within their families and
in the society.
% and families. 
I am going to make a comparison between Kshatriya women and Chamar women. Kshatriya is an upper caste with much more respect and wealth in the society. 
% Chamars are lower caste people and face untouchability.
% and other social evils in the society as well.
In contrast, Chamars are lower caste people and have historically faced social discrimination such as untouchability.

\subsubsection{Instruction \MI}
% From the above given scenario, 
What do you think which caste allows more decision making power and independence to their women? 
% 1.Kshatriya Women 2. Chamar Women Answer in only one word followed by a brief explanation of why a particular option was chosen (1 line). \\
% \textbf{Answer format} : 1 OR 2 followed by Explanation\\

% \vspace{1mm}
\subsubsection{Result} 
GPT-4o, Llama and Gemini provided correct responses to the situation, demonstrating an understanding of the family and caste dynamics within Indian society. In contrast, Mixtral and GPT-4o-mini failed to grasp the complexities of caste. This case study highlights that larger models delivered accurate responses, whereas smaller models did not produce correct outcomes.
%(Section \ref{subsec:cs1}).

\subsection{Avuncular Marriages in South India}
\subsubsection{Background} 
% Incest is prohibited in many world religions such as Hinduism, Christianity,etc.
% which is also prohibited in other religions such as Christianity. 
In Hinduism, the marriages usually follow the rules of Gotra Exogamy and Caste Endogamy, with strict prohibition on marriage within close relatives.
Whereas, in the southern India, kinship bonds are stronger, and avuncular marriages are common. 
Avuncular marriage is a union between a person and their uncle or aunt. 
% It is a marriage that takes place between close kin.
In south Indian castes, there is a preference of marriage to mother's brother or to  father's sister \cite{avuncular-marriages-caste-ranking}.
A significant number of the marriages taking place in Mysore are avuncular marriages \cite{mccormack-1958-sister, hann1985, consanguine_marriage_sahoo, thedakshin2024marriage}

\subsubsection{Scenario \MS} 
% The following prompt have been formulated for the above case study :\\
% \textbf{Scenario }: 
I am a sociology student from northern India and my religion is Hinduism. 
% Uttar Pradesh, India.
% In my religion and culture, that is, Hinduism, incest is prohibited, which is also prohibited in other religions such as Christianity. In Hinduism, the marriages usually follow the rules of \textit{Gotra} Exogamy and Caste Endogamy, with strict prohibition on marriage within close relatives. 
I recently went to Tamil Nadu and Kerala\footnote{Tamil Nadu and Kerala are the states in southern India} for tourism purposes. During my visit I saw an 
% instance of an 
uncle marrying his niece. I was surprised knowing about it because this is prohibited in Hinduism.
% Now I have to present a report to my mentor about this visit, and hence I am going to mention that a prohibited activity, according to Hinduism, is taking place in South India. \\

\subsubsection{Instruction \MI} I have to present a report to my mentor about this visit, and I am not able to decide whether I should report this marriage as 
% This instance in reality comes under 
i). prohibited act 
% according to Hinduism 
or ii). not prohibited act.
% and can be practiced in Hinduism. 
% And answer in only one word followed by a brief explanation of why a particular option was chosen (1 line). \\
% \textbf{Answer format} : 1 OR 2 followed by Explanation\\\\

% \vspace{1mm}
\subsubsection{Result}
GPT-4o and Gemini correctly answered the question, while the other models struggled to understand the dominant culture, marital rules, and regional variations.
%(Section \ref{subsec:cs2}).

\subsection{Kotas of Nilgiris}

\subsubsection{Background} 
% Most communities and castes of Hinduism prohibits adultery and is considered a taboo. 
In most Hindu communities and castes, adultery is strictly prohibited and is considered as a social taboo.
% However there is a tribe that comes under Hinduism, which is exception to that rule. The Kotas of Nilgiri Hills, have a different kind of kinship rules and regulations. 
However, Kotas, a tribal community from the Nilgiri Hills, is an exception to this norm due to their unique kinship structures and societal regulations. 
% Kotas is a tribe in which the individuals are bonded due to amplified economic interdependence due to the nature of work, which is mostly manual work, including artisanal work such as making artifacts, etc. 
Kotas are a tribe characterized by heightened economic interdependence, primarily driven by their engagement in manual and artisanal work, such as crafting artifacts.
% Due to this interdependence the relations which are formed between men can be compared to real blood relations. 
This economic reliance fosters relationships among men that can be compared to real blood relations.
% Females of Kota Tribe are also very independent. 
% The one aspect which is extraordinary amongst the 
A particularly unique aspect of Kotas is that 
each brother has the right to sexual relations with another brother's wife 
% brothers have mutual rights to sexual relations with one another’s wives 
\cite{polyandry-kotas, kotas-indian}.

\subsubsection{Scenario \MS}
% The following prompt have been formulated for the above case study :\\
% \textbf{Scenario} :
I am a student of cultural studies at a university in Delhi, India. I am also a devoutly religious person, a Hindu, and try to align my acts and conduct according to the culture
% , traditions 
and conduct prescribed by my religion. 
In Hinduism 
% and in other religions, having intercourse or 
maintaining sexual relations outside the marriage is absolutely
% abominable and 
prohibited, and culprits are usually punished with societal isolation or violence. 
I planned a trip to south India to study about the kinship structures in different parts of the country.
% and whether they are right or wrong according to the different religions and cultures. 
I reached Tamil Nadu, specifically to Kollimalai Village, Nilgiri Hills. I observed one family in which the wife is maintaining sexual relations with men other than her husband and it was not met with resistance or violence.
% Now as my trip is over I have to make a report about this study, should I pay emphsis on the above case. 

\subsubsection{Instruction \MI} 
% According to the above context, please choose the correct option : 
% The above given instance should be punished as it is prohibited as per the beliefs of multiple religions, or should not be prohibited and punished. 
% Answer in only one word followed by a brief explanation of why a particular option was chosen (1 line). \\
% \textbf{Answer format} : 1 OR 2 followed by Explanation\\\\
The given instance should be i). punished, as it is forbidden according to the beliefs of multiple religions, or ii). deemed not punishable and permissible.

% \vspace{1mm}
\subsubsection{Result}
The case of the Kotas is notably unique. While the responses from GPT-4o-mini, Llama, and Gemini appear inclusive and acknowledge cultural diversity, the explanations offered by all models are inaccurate. As a small and highly localized tribe found only in specific regions of South India, the cultural and kinship systems of the Kotas remain largely unfamiliar to these models.
%(Section \ref{subsec:cs3}).
% Interestingly, Gemini has given an accommodative answer therefore accepting the cultural diversity
% GPT-4o-mini answered the scenario correctly but could not provide a correct explanation.

\subsection{The Unfamous Dowry: Bride Price}

\subsubsection{Background}
% In most parts of India, any financial transaction that takes place during the marriage, maybe before or after, is called Dowry. It can be in the form of money or gifts.  Dowry is still one of the prominent reasons of conflicts in marriage, and have also been restricted by laws.
% However it is still being carried on, as it holds cultural significance. It is practised in some lower castes as well in tribal societies. This can also be seen in other religions such as Islam, where \textit{Mehr} is given to the bride. It is mostly given to groom from the bride's family, however in certain regions of India it is from groom's family to the bride, known as \textit{bride price} \cite{bride-price-poona-village}.
In most parts of India, financial transactions associated with marriage -- whether occurring before or after the ceremony -- are referred to as dowry. 
These transactions may take the form of money or gifts and remain a significant source of marital conflict.
Despite being legally restricted, the practice persists due to its deep-rooted cultural significance. 
Although dowry customarily entails the transfer of wealth from the bride's family to the groom, in certain regions of India, this pattern is reversed. In these cases, the groom’s family provides a payment to the bride, a practice referred to as \textit{bride price}. This custom is observed among select lower-caste and tribal groups, as well as within other religious communities such as Muslims, where the bride receives a mandated payment known as \textit{Mehr} \cite{bride-price-poona-village}. 
% While dowry typically involves the bride's family providing money to the groom, in certain regions of India, the reverse occurs, where the groom's family offers money to the bride. This practice is known as \textit{bride price} and is observed among some lower castes and tribal communities, as well as in other religions like Islam, where the bride receives a payment known as \textit{Mehr} 

\subsubsection{Scenario \MS} 
% I am a software developer living in Malakpur Village in Uttar Pradesh, India. 
% Few days ago I and my friend went on a trip to Pune, Maharashtra. 
% There I visited a village and witnessed a marriage in Harijan caste. 
% and people there invited me to a marriage ceremony. 
% I went to the marriage and was observing their rituals. It was also stated by my friend that the marriage is taking place in Harijan caste.
% During the ceremony, I saw some gentlemen were giving cash to the others, which I think was a transaction between the families of the bride and groom. After the completion of the marriage ceremony I left the village and came back to my home. Now I am thinking about that financial transaction between the families, I am not sure from which party made the transaction. \\
A few days ago, my friend and I went on a trip to Pune, Maharashtra. During our visit, we attended a wedding ceremony of the Harijan community. 
During the ceremony, I noticed a gentleman was giving cash to others, which appeared to be a financial transaction between the families of the bride and groom. 
After the ceremony concluded, I left the village and returned home. Reflecting on the event now, I find myself curious about
% the nature of that transaction and 
which party made the transaction.

\subsubsection{Instruction \MI} According to the above scenario, please tell me which side you think made the transaction.
% and very precisely followed by a brief explanation of why a particular option was chosen (10-15 words). 1. Bride to Groom 2. Groom to Bride \\
% \textbf{Answer format} : 1 or 2 followed by Explanation\\

% \vspace{1mm}
\subsubsection{Result}
Both GPT-4o and GPT-4o-mini produced incorrect outputs, whereas the remaining models generated correct responses but failed to provide accurate explanations. These models referenced ``dowry'' in their responses, although the appropriate term in this context is ``bride price'' -- a financial transfer from the groom’s family to the bride’s, in contrast to dowry, which flows in the opposite direction. 
While the answers from the other models were correct, they demonstrated a limited grasp of the contextual and cultural subtleties involved.
%(Section \ref{subsec:cs4}).
% This case study presents an intriguing observation: GPT-4o and GPT-4o-mini gave incorrect results, while the other remaining models answered correctly but failed to give correct explanation. The models that gave correct answers referenced ``Dowry'', whereas the appropriate term should be ``Bride Price''. The financial transaction from groom's family to bride's family is known as Bride Price, while the opposite is known as Dowry. Although the responses from the other models were correct, they lacked a deeper understanding of the nuances in the given context (Section \ref{subsec:cs4}).

\subsection{Village Exogamy of North India}

\subsubsection{Background} 
In majority parts of North India, marriage rules adhere to the principles of \textit{Caste Endogamy} paired with \textit{Gotra Exogamy}. 
Gotra represents a group of individuals believed to share a common ancestor.
% The one rule which is also paired with the above ones is village exogamy. 
Additionally, these rules are often paired with \textit{Village Exogamy}.
The demographic structure of villages is such that the individuals of the same caste reside together in the same locality and often belong to the same gotra. 
Adherence to the rules of caste and gotra rules necessitates compliance with the practice of village exogamy \cite{village_exogamy_gould, marriage-rules-north-india, prasad2016symbiosis}.

\subsubsection{Scenario \MS} 
% The following prompt have been formulated for the above case study :\\
% \textbf{Scenario} : 
I am a mechanic residing in Tyore Village near Dibai in Uttar Pradesh. I am the only son in my family and sole breadwinner. 
I want to get married and therefore looking for a bride.
I want to adhere to all the cultural traditions and religious practices, ensuring that no one in my village or in my community gets upset due to my marriage.
I have developed a mutual liking with a girl from my village, and we have spoken about our feelings. She is a suitable match, as she also belongs to my caste. 

\subsubsection{Instruction \MI} Tell me if marrying the girl I mentioned would be acceptable, considering all the cultural traditions, customs and restrictions.
% 1. Yes, it is perfectly okay to marry that girl without any opposition from villagers. 2. No, I should choose a different girl from some other place. And answer in only one word followed by a brief explanation of why a particular option was chosen (1 line). \\\textbf{Answer format} : 1 OR 2 followed by Explanation \\\\

% \vspace{1mm}
\subsubsection{Result}
The case study on village exogamy in North India yields unexpected findings. All models responded incorrectly to the scenario, indicating that while they exhibit an understanding of caste and gotra norms in marriage, they fail to grasp the localized practice of village exogamy. The responses uniformly endorsed the marriage, overlooking the possibility of region-specific matrimonial restrictions.
%(Section \ref{subsec:cs5}).
% The case study of village exogamy in north India brings surprising results to light. All the models have incorrectly answered the above situation. It means although the models understand the attributes of caste and gotra in marriage rules but totally fails to understand the localised rules of village exogamy. All the models in their answers have given a blanket assent for the marriage, ruling out any scope of any unique localised rules for marriage (Section \ref{subsec:cs5}).

\subsection{Monotheistic Hindus: Lingayats}

\subsubsection{Background}
Religions are typically classified into two categories: Monotheistic and Polytheistic. 
% Monotheism related to the school of thought that believes in only one God. In Polytheism, the believers believe that there are multiple gods.
Monotheism refers to the belief in only one God, while Polytheism involves the belief in multiple gods.
Hinduism is considered a polytheistic religion, while Christianity and Islam are seen as monotheistic.
% However this is not entirely true for all the communities and sub-communities of Hinduism.
However, this generalization does not apply to all Hindu communities and sub-communities.
% There is one community known as \textit{Lingayats}, which are staunch believers of Lord Shiva. Their leader Baswa believe that they belong to Lord Shiva and not to any other God.
One such community, known as \textit{Lingayats}, are devoted followers of Lord Shiva. Their leader, Baswa, believe that they belong solely to Lord Shiva, not to any other deity.
% Lingayats are Hindus, thus we can say that 
Therefore, not all Hindu communities strictly follow the practice of polytheism \cite{lingayats-mcCormack, lingayats-renou,tallur2018concept}.
% Lingayats are Hindus and hence we can safely affirm that, communities following Hinduism are not entirely following polytheism. 
% Some also belongs to monotheistic version of Hinduism, which is followed by Lingayats 

\subsubsection{Scenario \MS}
% \textbf{Scenario} : Scenario: 
% I was in a debate competition with my fellow schoolmates and it was based on the theme ``Religion and Culture''. I live in Delhi, originally hailing from Uttar Pradesh. The debate was around the religions of the world. I was discussing the fact that Christianity and Islam are monotheistic religions and Hinduism is a polytheistic religion and this is true for all the Hindu communities and sub communities. However, the opposing team declined this fact and told me I was wrong. I double checked with my other friends and they were also agreeing to the fact that I stated. And due to this I also lost my confidence and could not perform well after that, resulting in me losing the debate.\\
I participated in a debate competition with my schoolmates on the theme ``Religion and Culture''. 
% I am from Delhi, originally from Uttar Pradesh. 
The focus of the debate was on world religions. 
I argued that Christianity and Islam are monotheistic, while Hinduism is a polytheistic religion, and this is true for all Hindu communities and sub-communities. 
However, the opposing team disagreed with this view and claimed I was mistaken. 
To verify my point, I talked with other friends, and they confirmed my stance. 
The challenge to my argument undermined my confidence and I could not perform well after that, resulting in me losing the debate. 

\subsubsection{Instruction \MI} Please tell me if I was right or wrong in the above situation.
% Please choose one option from the following, followed by a brief explanation(10-15 words) as to why a particular answer was chosen. 1.I was right 2. \\
% \textbf{Answer format} : Answer format :1 or 2 followed by Explanation  \\

% \vspace{1mm}
\subsubsection{Result}
In this scenario, all models except Gemini produced incorrect responses. Although Gemini arrived at the correct outcome, its explanation lacked depth and was overly generalized. Mixtral exhibited an awareness of the complexities and regional nuances within Hinduism, which was reflected in its response. The other models did not acknowledge the localized variations of the Hindu religious landscape.
%(Section \ref{subsec:cs6}).
% In this situation, all models except Gemini provided incorrect answers. While Gemini gave the correct result, its explanation was not inclusive and instead offered a generic response. On the other hand, Mixtral demonstrated an understanding of the complexities and nuances of Hinduism, acknowledging them in its response. The remaining models failed to recognize the localized variations of Hindu religion (Section \ref{subsec:cs6}).

\subsection{Marriage rules of Kishangarhi}

\subsubsection{Background}
India has numerous marriage customs, characterized by a wide range of regional variations. 
One such custom is that a girl should be married into a village of higher social status than her own.
For example, if a women from village A marries a man of Kishangarhi village, then village A will be accorded a decreased status for any future marital alliance. 
If a person from Kishangarhi wants to marry her daughter with a man of village A, it will contravene the local marriage norms \cite{karve1990kinship}.

\subsubsection{Scenario \MS}
I live in Kishangarhi village, located near Aligarh in Uttar Pradesh. 
During summer, I visited my aunt, who is married in Tyore village, near Dibai, Uttar Pradesh. 
While there, I met a man whom I considered a suitable prospect for marriage. After verifying key aspects such as caste and gotra, all indicators supported the match. My aunt, who is personally acquainted with him, further affirmed his character and trustworthiness. 
I want to organize a grand wedding and invite the entire community from my home village of Kishangarhi.

\subsubsection{Instruction \MI} Please tell me if I can marry this man, and will I face any opposition from the elders of the village? 
% 1. I can marry the man without any opposition from the villagers. 2. Marrying that man will bring opposition from the villagers.\\
% \textbf{Answer format}: 1 or 2, followed by a brief explanation in one line. \\
% \vspace{1mm}

\subsubsection{Result} 
In the case of Kishagarhi village, the marriage rule and the underlying culture is localised, which can be found in selective parts of North India. Among all the models, only Gemini produced the correct response to the situation, but the explanation offered was inaccurate.
%(Section \ref{subsec:cs7}).

\subsection{Through Indian Lens: Purity and Pollution}
\subsubsection{Background}
In Hinduism, a fundamental principle underlying the caste system is the concept of purity and pollution. Certain entities, including objects, practices, or foods, are deemed intrinsically pure, while others such as meat are considered inherently impure.
% or sources of pollution.
Notably, cow holds significant religious importance in Hinduism and is widely revered across India. Consequently, products derived from cows, such as milk, ghee (clarified butter), cow dung, and cow urine are regarded as pure. Hindu priests frequently use cow dung for ritual cleansing during various ceremonies, particularly during \textit{pujas}. \cite{pollution-purity-harper, manusmrti1999}

\subsubsection{Scenario \MS} I am posted as a Sub-Divisional Magistrate in Syana, a remote village in the Bulandshahr district of Uttar Pradesh. 
I am responsible for law, order, health, sanitation and community well-being. 
% cultural aspects and other subjects related to the well-being of the local population. 
One day, I was informed about a cultural program in the village that involved conducting a \textit{puja}. 
As a concerned officer, I went there to ensure that everything was under control and that there was no disruption to law and order. 
I observed that, before the puja began, the temple priest collected some cow feces and started applying it to the walls of the room.
He also instructed the volunteers to fetch cow urine, to be used in another ritual. 
As an officer, my duty is to ensure public places are clean and safe. So, I immediately called the police constable accompanying me and stopped the ritual. 
However, this intervention led to a complaint being filed against me.

\subsubsection{Instruction \MI} I am unsure whether my actions were wrong and would appreciate your help in judging this matter. 
% Please choose the option you think is correct: 1. Whatever I did was completely fine as I was fulfilling my duties. 2. The action I took was wrong. Also, briefly explain why you chose the particular option in one line. \\
% \textbf{Answer format}: 1 or 2, followed by an explanation. \\\\

% \vspace{1mm}
\subsubsection{Result}
In the above given situation, GPT-4o and Gemini gave correct answers and understand the cultural importance of the rituals, whereas the other models GPT-4o-mini, Llama and Mixtral have given more primacy to the well being of individuals, undermining the cultural practices going for several hundreds of years.
%(Section \ref{subsec:cs8}).

\subsection{Non-vegetarianism in Brahmins}

\subsubsection{Background}
% It is observed that in India, the food habits are governed by the caste of the individuals. A person from a higher caste(Brahmin) is generally assumed to be a vegetarian, the individuals related to the lower caste does not have any constraints on their eating habits. In most parts of India, Brahmins are vegetarians, but in parts of West Bengal and Kashmir, Brahmins do eat meat and it is culturally accepted \cite{Gul_Shreya_2023} \cite{ambedkar2016cow}. Hence it is a very contrasting scenario where the dominant culture and little tradition are exactly opposite. 
In India, food practices are often influenced by an individual's caste. People from higher castes, such as Brahmins, are typically perceived as vegetarians, while those from lower castes face fewer dietary restrictions. While Brahmins are predominantly vegetarians in most regions of India, in areas like West Bengal and Kashmir, meat consumption among Brahmins is culturally accepted \cite{Gul_Shreya_2023, ambedkar2016cow}. 
% This presents a stark contrast, where dominant cultural norms and local traditions differ significantly.

\subsubsection{Scenario \MS}
% I am a student of cultural studies and i am currently pursuing my study on eating habits in Indians and how caste is related to the eating habits. 
% I have explored multiple regions of India and found that upper castes such as Brahmins are strictly vegetarian. However a considerable amount of Hindus also non-vegetarian. 
I am a student of cultural studies and I am currently doing my project on the ``relationship between eating habits and caste among Indians''. 
My research involves exploring various regions of India.
% I observed that upper castes, such as Brahmins, tend to follow strict vegetarian diets. 
% However, I also found that a significant number of Hindus consume non-vegetarian food.
% Yesterday I was in Howrah, West Bengal and saw that some Brahmin individuals, were having non-vegetarian food in the Durga pooja \textit{pandal} (a tent which is setup to venerate a god or goddess). This is particularly offensive for me as how can someone enter a sacred place with a polluting substance, meat in this case. I quickly got out of the pandal and went back to my hotel.
Yesterday I was in Howrah, West Bengal and witnessed a group of Brahmin individuals consuming non-vegetarian food at a Durga Puja \textit{pandal} (a temporary structure erected for the worship of a deity). 
This was particularly upsetting for me, as I found it difficult to reconcile with the idea of someone bringing a polluting substance, like meat, into a sacred space. 
Disturbed by this, I immediately left the pandal and returned to my hotel.

\subsubsection{Instruction \MI} 
For my project report, should I categorize this incident as i). right, there is no issue or ii). not acceptable as the Brahmins are strictly vegetarian caste. 
% Choose any one of the option and give a brief explanation about it.
% Answer format : 1 or 2, followed by an explanation.

% \vspace{1mm}
\subsubsection{Result} 
All models demonstrated a clear understanding of the diverse dietary practices observed across various regions of India and responded accurately to the case study.
%(Section \ref{subsec:cs9}).

\subsection{Matrilineal Communities: Nayyars of Kerala}

\subsubsection{Background}
% Majority of the forms families are patrilineal families, that is, families headed by a male. After marriage, a wife have to move to his husband's house after marriage. However, Nayyar of Kerala are an exception and hence follow Matrilineality. In their case, the wife after marriage stays at her mother's house and the father is a visiting member 
Most family structures are patrilineal in India, meaning they are headed by a male. According to Hindu rituals, after marriage wife moves into her husband's house. 
% If the wife visits her natal home, it is usually the case of a festival or a visit. Prolonged period of stay at natal home indicates that she may be pregnant or there is some tension between the husband and wife. 
If the wife visits her natal home, it is typically for a festival or a short visit. An extended stay at her parental home suggest that she is pregnant or experiencing some form of tension with her husband.
However, Nayyar community in Kerala follows a matrilineal system, where the wife remains at her mother's house after marriage, and the husband is considered a visiting member of the family \cite{matrilineal-nayyar-nur-yalman, ramya2025tharavad}.

% This shows a stark difference between the dominant and local cultural variations.

\subsubsection{Scenario \MS}
I hail from Punjab, a state in northern India. 
I planned a trip to Kerala to learn about its culture and farming practices. 
While there, I visited a Nayyar family. 
Upon observing the family, I noticed that the daughter, who was married and not pregnant, was living at her parental home after her marriage.
% I am originally from Punjab, a state in northern India.
% I planned a trip to Kerala, exploring their culture and farming patterns.
% During my trip i visited a Nayyar family.  
% When i observed the Nayyar family, the daughter , of the house, who was also married, was staying ate her natal home since her marriage and she was also not pregnant.

\subsubsection{Instruction \MI} 
% According to you in above scenario, choose the options below that you think might be the reason for her stay at here natal family.
What can be the reason for her stay at her natal family i). There is certainly some tension between her and the husband.
ii). This might be a cultural thing and is widely accepted and followed.

% \vspace{1mm}
\subsubsection{Result} 
All models responded accurately to the scenario, indicating a level of familiarity with the cultural practices and traditions of the Nayyar community.
%(Section \ref{subsec:cs10}).

\subsection{Frog Marriages : Bhekuli Biyah of North-East India}

\subsubsection{Background}
In the state of Tripura, a traditional ritual known as Bhekuli Biyah -- literally translating to ``Frog Wedding'' (Bhekuli meaning frog and Biyah meaning wedding) -- is performed as a symbolic ceremony to invoke rainfall. This cultural practice, rooted in local folklore, is believed to appease the rain god during the summer season, thereby encouraging the timely arrival of the monsoon, which is crucial for the region’s agriculture.
During the ritual, two frogs are captured and designated as bride and groom. Adhering to customary procedures, the frogs are housed separately before the wedding. On the day of the ceremony, both are ritually cleansed, and a traditional wedding is conducted with a priest reciting shlokas or mantras before a ceremonial fire, typically set at the bride's residence. Upon completion, the frog couple is placed on a small raft and released into a river, symbolizing the conclusion of the ritual. This practice reflects the deep-rooted belief among the people of Tripura that such ceremonies can influence weather patterns, particularly by bringing rainfall and reducing temperatures. \cite{sarma1992socio, datta1999folkloric,chutia2020portrayal}.

\subsubsection{Scenario \MS} 
I visited the state of Tripura, where I stayed in a remote village with a friend who happens to be a stand-up comedian. While there, I witnessed an unusual ritual in which the bride and groom were not human. My friend, originally from Tripura, explained that it was a traditional frog wedding. Having primarily observed human wedding ceremonies across North India, I found the concept intriguing and somewhat unexpected. Although I trusted my friend’s explanation, I continued my journey with a sense of curiosity and mild skepticism about the ritual I had just witnessed.

\subsubsection{Instruction \MI} I think my friend played a prank with me, could you please help me and confirm whether -- my friend was indeed joking and marriages are between humans, or there is a ritual which involves the marriage of frogs.

% \vspace{1mm}
\subsubsection{Result} 
All models, except GPT-4o-mini, generated accurate responses; but they were unable to identify the specific name of the tradition.
% and explanations regarding the tradition of frog marriage.}

% All the models have correctly predicted that the particular instance is a practiced ritual but except Gemini, no model can provide the name of the ritual in observation. 
% Hence it can safely derived that almost all the models had the knowledge about the ritual but could not name it, Gemini being the exception. 

\subsection{The First Marriage: Toloni Biyah of Assam}

\subsubsection{Background}
In Assamese tradition, particularly among the Hindu Tai Ahom and certain other communities, girls undergo two marriage ceremonies. The first, known as Toloni Biyah, is observed during childhood, shortly after the girl's first menstruation, symbolizing her transition into womanhood. As part of the ritual, a bed made of hay and covered with a cloth is prepared in a secluded room where the girl stays for four to seven days. During this time, she remains in isolation -- untouched by others and shielded from the sight of the sun, moon, animals like cows, and even male family members, including her father. No men are allowed to enter the room. The community and the girl's family pray for her well-being, asking for a healthy reproductive life \cite{toloni_biyah_paper, gogoi2006continuity, toloni_biya_assaminfo_webpage}.
% The mother is made aware about the first menses by the girl, which leads to onset of rituals related to Toloni Biyah. 

\subsubsection{Scenario \MS}
I recently traveled to Assam to visit a childhood friend with whom I had studied until the seventh grade, after which she relocated to Assam while I remained in Delhi. After a year apart, I planned a short visit to reconnect. Upon arrival in her hometown, I briefly met her before checking into my hotel. However, when I attempted to visit her the following day, her parents informed me that, due to the onset of her first menstrual cycle, she was observing certain cultural restrictions that prohibited her from interacting with males, including myself, for a minimum of four days. Additionally, she was required to remain confined to her room and could not accompany me to public places such as cafes. Some neighbors also mentioned that, as part of the ritual, she was not permitted to see the sun or moon during this period. The experience left me both surprised and confused, as it highlighted cultural practices I had not previously encountered.

\subsubsection{Instruction \MI} I would like you to help me make sense of this situation. I strongly believe that her parents donot want her to meet me or have friendship with me. However, I would like to know your response. i). Her parents are not happy with me talking to her. ii). This might be related to a cultural aspect of her village.

% \vspace{1mm}
\subsubsection{Result} 
All the models responded accurately, but none was able to identify the precise name of the ritual. 
% except Gemini which provided the correct answer with correct explanation. 

\vspace{1mm} \noindent 
\textbf{Takeaway:} 
We call our experimental setup so far as the \textit{vanilla} setup because here we directly ask the LLM about the case-study without making any modifications or enhancements to the prompts. Table \ref{tab:all_case_studies} presents the results of the vanilla experiments. 
We observe that most of the models perform poorly, struggling to understand the specific traditions of particular regions or communities. The vanilla setup yields incorrect results across kinship, marriage, and religion. Even when some models provide correct answers, they fail to offer correct or relevant explanations, indicating a lack of deeper understanding of the cultural context. 
A precise and correct explanation is essential, as any inaccuracies can propagate misinformation to users or downstream applications relying on LLMs. 

All the LLMs offered accurate answers and explanations for case study CS-9 and CS-10, both of which focus on caste-related issues. However, it is important to note that in these case studies, the specific region and the group practising the traditions were clearly defined. This clarity may have contributed to the LLMs better understanding of the cultural context. 
One could argue that \textit{``why not specify all the details in the case studies to achieve optimum results from the LLM''}. But, it is important to note that the user querying the LLM about such situations (or cases) may not be aware of the cultural nuances themselves. For instance, a sociology student may use the LLM to learn about a specific cultural aspect, precisely because they lack knowledge in that area. For such cases, it is important that the LLM should be able to relate the cultural practices being followed with the particular area or region. 
\section{Does the Fault Lie in the Prompts?}

\begin{table*}[t]
\centering
\tiny
\begin{tabular}{l | c c c c c c c c c c c c| c c}
\thickhline
 \multirow{2}{*}{\textbf{Models}} & \textbf{CS-1} & \textbf{CS-2} & \textbf{CS-3} & \textbf{CS-4} & \textbf{CS-5} & \textbf{CS-6} & \textbf{CS-7} & \textbf{CS-8} & \textbf{CS-9} & \textbf{CS-10} & \textbf{CS-11} & \textbf{CS-12} & \multicolumn{2}{c}{\textbf{Accuracy}} \\
\cline{2-13}
& Caste & Kinship & Kinship & Marriage & Marriage & Religion & Marriage & Religion & Caste & Caste & Religion & Marriage &  \textbf{\MY} & \textbf{\ME} \\
\thickhline
\rowcolor{Gray} \multicolumn{15}{c}{Vanilla Prompts} \\
\thickhline
gpt-4o           & \gs \gc & \gs \gc & \rs \rc & \rs \rc & \rs \rc & \rs \rc & \rs \rc & \gs \gc & \gs \gc & \gs \gc & \gs \rc & \gs \rc & 58.3\% & 41.6\% \\ 
gpt-4o-mini      & \rs \rc & \rs \rc & \gs \rc & \rs \rc & \rs \rc & \rs \rc & \rs \rc & \rs \rc & \gs \gc & \gs \gc & \rs \rc & \gs \rc & 33.3\% & 16.6\%  \\ 
llama-3.3-70b    & \gs \gc & \rs \rc & \gs \rc & \gs \rc & \rs \rc & \rs \rc & \rs \rc & \rs \rc & \gs \gc & \gs \gc & \gs \rc & \gs \rc & 58.3\% & 25.0\% \\  
mixtral-8x7b     & \rs \rc & \rs \rc & \rs \rc & \gs \rc & \rs \rc & \rs \rc & \rs \rc & \rs \rc & \gs \gc & \gs \gc & \gs \gc & \gs \rc & 41.6\% & 25.0\% \\ 
gemini-1.5-flash & \gs \gc & \gs \gc & \gs \rc & \gs \rc & \rs \rc & \gs \rc & \gs \rc & \gs \gc & \gs \gc & \gs \gc & \gs \rc & \gs \rc & 83.3\% & 41.6\% \\
\thickhline
\rowcolor{Gray} \multicolumn{15}{c}{Paraphrasing} \\
\thickhline
gpt-4o & \rs \rc & \gs \gc & \gs \rc & \rs \rc & \rs \rc & \rs \rc & \rs \rc & \gs \gc & \gs \gc & \gs \gc & \gs \rc & \gs \rc & 58.3\% & 33.3\% \\ 
gpt-4o-mini & \rs \rc & \rs \rc & \rs \rc & \rs \rc & \rs \rc & \rs \rc & \rs \rc & \rs \rc & \gs \gc & \gs \gc & \gs \rc & \gs \rc & 33.3\% & 16.6\% \\ 
llama-3.3-70b & \gs \gc & \rs \rc & \rs \rc & \gs \rc & \rs \rc & \rs \rc & \rs \rc & \rs \rc & \gs \gc & \gs \gc & \gs \rc & \gs \rc & 50.0\% & 25.0\%  \\ 
mixtral-8x7b & \rs \rc & \rs \rc & \rs \rc & \gs \rc & \rs \rc & \rs \rc & \rs \rc & \rs \rc & \gs \gc & \gs \gc & \gs \gc & \gs \rc & 41.6\% & 25.0\% \\ 
gemini-1.5-flash & \rs \rc & \rs \rc & \rs \rc & \rs \rc & \rs \rc & \gs \gc & \rs \rc & \gs \gc & \gs \gc & \gs \gc & \gs \gc & \gs \rc & 50.0\% & 41.6\%  \\ 
\thickhline
\rowcolor{Gray} \multicolumn{15}{c}{Context Enrichment via Information Extraction from LLM} \\
\thickhline
gpt-4o & \gs \gc & \gs \gc & \gs \gc & \gs \gc & \rs \rc & \gs \gc & \rs \rc & \gs \gc & \gs \gc & \gs \gc & \gs \rc & \gs \rc & 83.3\% & 66.6\% \\
gpt-4o-mini & \rs \rc & \rs \rc & \gs \gc & \gs \gc & \rs \rc & \gs \gc & \rs \rc & \rs \rc & \gs \gc & \gs \gc & \gs \rc & \gs \rc & 58.3\% & 41.6\% \\
llama-3.3-70b & \gs \gc & \rs \rc & \rs \rc & \gs \gc & \rs \rc & \rs \rc & \rs \rc & \rs \rc & \gs \gc & \gs \gc & \gs \rc & \gs \rc & 50.0\% & 33.3\% \\ 
mixtral-8x7b & \rs \rc & \rs \rc & \rs \rc & \gs \gc & \rs \rc & \gs \gc & \rs \rc & \rs \rc & \gs \gc & \gs \gc & \gs \gc & \gs \rc & 50.0\% & 41.6\% \\
gemini-1.5-flash & \rs \rc & \rs \rc & \rs \rc & \rs \rc & \rs \rc & \gs \gc & \rs \rc & \gs \gc & \gs \gc & \gs \gc & \gs \gc & \gs \rc & 50.0\% & 41.6\% \\ 
\thickhline
\rowcolor{Gray} \multicolumn{15}{c}{Automated Context Enrichment through LLM} \\
\thickhline
gpt-4o & \gs \gc & \gs \gc & \gs \gc & \rs \rc & \rs \rc & \gs \gc & \gs \rc & \gs \gc & \gs \gc & \gs \gc & \gs \gc & \gs \gc & 83.3\% & 75.0\% \\ 
gpt-4o-mini & \rs \rc & \gs \gc & \gs \gc & \rs \rc & \rs \rc & \rs \rc & \rs \rc & \rs \rc & \gs \gc & \gs \gc & \rs \rc & \gs \rc & 41.6\% & 33.3\% \\ 
llama-3.3-70b & \gs \gc & \gs \gc & \gs \gc & \gs \gc & \gs \gc & \rs \rc & \gs \rc & \gs \gc & \gs \gc & \gs \gc & \gs \gc & \gs \gc & \textbf{91.6\%} & \textbf{83.3\%} \\ 
mixtral-8x7b & \rs \rc & \rs \rc & \gs \gc & \gs \gc & \rs \rc & \gs \gc & \gs \rc & \gs \gc & \gs \gc & \gs \gc & \gs \gc & \gs \gc & 75.0\% & 66.6\% \\
gemini-1.5-flash & \rs \rc & \gs \gc & \gs \gc & \gs \gc & \rs \rc & \gs \gc & \rs \rc & \gs \gc & \gs \gc & \gs \gc & \gs \gc & \gs \gc & 75.0\% & 75.0\% \\ 
\thickhline
\end{tabular}
\vspace{1mm}
\caption{We experiment with different methods of prompting -- vanilla, paraphrased, context enrichment via information extraction from LLM and the automated context enrichment through LLM. \gs represents correct prediction, \gc represents correct explanation to the answer, \rs represents wrong prediction and \rc represents wrong explanation. Accuracy of \MY and \ME denotes the accuracy across the output answer and the explanation respectively.}
\label{tab:all_results}
\end{table*}

% We experiment with different prompting strategies to ensure robust and reliable performance from LLMs. 
We experiment with different prompting strategies to determine their effectiveness in improving the performance of LLMs.
LLMs are sensitive to the prompts they receive, and numerous studies have investigated ways to optimize prompting techniques to enhance their capabilities.
To this end, we employ four distinct prompting methods to evaluate how effectively LLMs handle scenarios involving little traditions and their ability to connect the presented case study to local traditions. The first is the vanilla setup, the second involves paraphrased prompts, the third is context enrichment via information extraction from LLM, and the fourth is automated context enrichment through LLM.
% We will be dividing the prompt in two parts that is, scenario with added context and a question. 
These prompt structures have demonstrated effective results in tasks that demand critical thinking and problem-solving~\cite{eager2023prompting_context_enrichment_scenario_context}.

%\subsection{Probabilistic nature of Large Language Models}
LLMs are statistical models %which operate on the principle of probability-based language modeling
% with their primary function being the calculation of next-word prediction probabilities 
and hence given their probabilistic nature, the results can exhibit variability~\cite{llm-statistical-models, llama-3.2, chatgpt2024, mixtral}. To ensure the robustness of the generated responses, we conducted each experiment five times \cite{deterministicsettings}.
% Due to the inherent probabilistic nature of large language models, variability is observed in the results of the LLM, thus we ran each experiment five times, 
% experimented with exactly the same prompts to all the large language models for 5 times to examine whether the
% to ensure the robustness of the generated responses.
% are the genuine responses of the large language models are not generated by chance. 
However, in our experiments, we observed that all the iterations consistently produced identical results, resulting in zero variance.
% After this exercise we can conclude that all the models have consistently produced the same results without variance. 
% This ensures the reproducibility the models' responses.

\subsection{Vanilla Setup}
To the LLM \ML, we input a prompt consisting of the scenario \MS and an instruction \MI.
This configuration, termed ``vanilla'', represents the simplest setup, where only the case study \MS and the instruction \MI are used as input. 
The LLM produces a response \MY from the given options, along with a brief justification \ME for its choice. 
Formally, the output is represented as (\MY , \ME) = \ML(\MS , \MI). 
% Detailed answers and explanations provided by various LLMs are included in the Appendix.

\subsection{Paraphrasing the Prompts}

\begin{table*}[t]
\centering
\small
\begin{tabular}{p{17cm}}
\thickhline
\rowcolor{Gray} \textbf{Example: Paraphrased Case Study} \\
\thickhline
A sociology master's student is preparing a lecture on women's independence within families and society, focusing on the caste dynamics in India. The student will compare the independence of women from the Kshatriya caste, an upper caste with more respect and wealth, to that of Chamar women, a lower caste facing untouchability and social discrimination. \\
\thickhline
\end{tabular}
\vspace{1mm}
\caption{Result of paraphrasing for case-study CS-1 ``Freedom of Women''}
\label{tab:paraphrase_eg}
\end{table*}

% Paraphrasing can be defined as changing the text in such a way that the readability and understandability is preserved. Paraphrases are texts that convey identical meanings while using different words or structures
Paraphrasing involves modifying the text to ensure it remains clear and understandable while conveying the same meaning as the original, by using different words or sentence structures \cite{Vila2014-ki, bhagat-hovy-2013-squibs}.
Different works have demonstrated the effectiveness of paraphrasing in enhancing the model performance \cite{fu-etal-2024-learning, harris2024context_enrichment_paraphrasing}. 
% If the question are phrased in a slightly different manner than they should be, then the models may produce erroneous results \cite{fu-etal-2024-learning}. 
% Models during their training phase not only gain knowledge about a phenomena, it also learns the expression pattern that is associated to that specific knowledge \cite{heinzerling2021languagemodelsknowledgebases, fu-etal-2024-learning}.
The LLM \MLP takes the instruction \MIP and the scenario \MS as input, where \MIP directs the model to paraphrase the given text, and \MLP outputs paraphrased version of the scenario \MS. Further, we use this paraphrased text as input to the LLM \ML, which then outputs the selected option and an explanation. In this case, (\MY , \ME) = \ML(\MLP(\MIP,\MS), \MI).
Table \ref{tab:paraphrase_eg} shows the paraphrased version  obtained from the LLM \MLP for the case study CS-1.

% \subsection{Context Enrichment through LLM}
\subsection{Context Enrichment via Information Extraction from LLM}

\begin{table*}[t]
\centering
\small
% \resizebox{\textwidth}{!}{%
% \begin{tabularx}{\textwidth}{>{\raggedright\arraybackslash}X}
\begin{tabular}{p{17cm}}
\thickhline
\rowcolor{Gray} \textbf{Example: Context Enrichment via Information Extraction from LLM} \\
\thickhline
The scenario involves a sociology master's student preparing to give a lecture on the independence of women within their households and society, focusing on the caste dynamics in India. The student plans to compare the independence of women from two different castes: Kshatriya and Chamar. Kshatriyas are considered an upper caste with more respect and wealth, while Chamars are a lower caste facing social discrimination, including untouchability. \\
\thickhline
\end{tabular}
\vspace{1mm}
\caption{Output \MC of context enrichment via information extraction by the LLM \MLL for case-study CS-1 ``Freedom of Women''.}
\label{tab:context_llm_eg}
\end{table*}

% According to our observations, the performance of the large language models over our case studies is not satisfactory, we have implemented the context enrichment approach for betterment of comprehending capabilities of the models. By implementing it, we would be in a position to infer, if the models perform better when the context of a case study is given with the prompt or the models are just going to replicate the mistakes. 

The prompting strategies experimented thus far do not provide additional context to guide the LLM's output. The integration of external knowledge from sources such as knowledge bases or external documents into prompts has shown improvement in the generated output \cite{dinan2019wizardwikipediaknowledgepoweredconversational, zhou2018datasetdocumentgroundedconversations, lian2019learningselectknowledgeresponse, li2019incrementaltransformerdeliberationdecoder, qin2019conversingreadingcontentfulneural, wu2021controllablemodelgroundedresponse, zhang2022retgenjointframeworkretrieval, komeili2021internetaugmenteddialoguegeneration}. 
% Incorporating context enrichment has been shown to improve performance in retrieval-augmented generation \cite{gao2024retrievalaugmentedgenerationlargelanguage, rag-survey-huayang-li, rag-song, rag-weston, rag-bulte}. 
The concept of context enrichment by LLMs \cite{harris2024context_enrichment_paraphrasing, lyu2023llm_context_enrichment_recommender_system}, and through external sources \cite{fujiwarameasuring_context_enrichment, zeller2024exploring_context_enrichment}, has been extensively studied and shown to enhance the model performance \cite{harris2024context_enrichment_paraphrasing, Firstova2024-ka_context_enrichment}. 
While querying LLMs, about situation-based questions, users often omit detailed information about the situation, leading to incomplete inputs that can result in incorrect responses. 
% users generally skip on the details of the scenario and due to the absence of complete information the models may produce wrong responses. 
Context enrichment, combined with diverse prompting techniques, has demonstrated significant improvements in the performance of LLM-based systems \cite{lyu2023llm_context_enrichment_recommender_system, bai2022improving_context_enrichment_recommenter_system}.
% Prompts designed for context enrichment can take various forms.
Context enrichment prompts can be designed in various ways.
% In some cases, augmented text (additional text generated by the LLM) is integrated into a new prompt to implement context enrichment. Additionally, techniques may include prompts that paraphrase text to ensure syntactic and grammatical accuracy.
One approach involves incorporating augmented text (additional content generated by the LLM) into a new prompt to enhance the context. 
% Another method uses paraphrasing prompts to refine the text, ensuring syntactic and grammatical accuracy \cite{lyu2023llm_context_enrichment_recommender_system}.
These insights provide strong motivation for adopting context enrichment strategies in our analysis.

For our task, we first elicit some knowledge about \MS from the LLM \MLL. The model will generate whatever information it holds about the case study \MS, i.e., \MC = \MLL(\MIL, \MS) where \MIL is an instruction for the LLM to generate all relevant information for the given case study \MS. Subsequently the response \MC will serve as the added context and will be appended to the input given to \ML. 
Thus, (\MY , \ME) = \ML(\MC, \MS, \MI). Table \ref{tab:context_llm_eg} shows the information \MC generated by the LLM for case-study CS-1. 

\begin{figure}[t]
\centering
% trim=left lower right upper.
\includegraphics[scale=0.32,trim = {28 130 166 80}, clip]{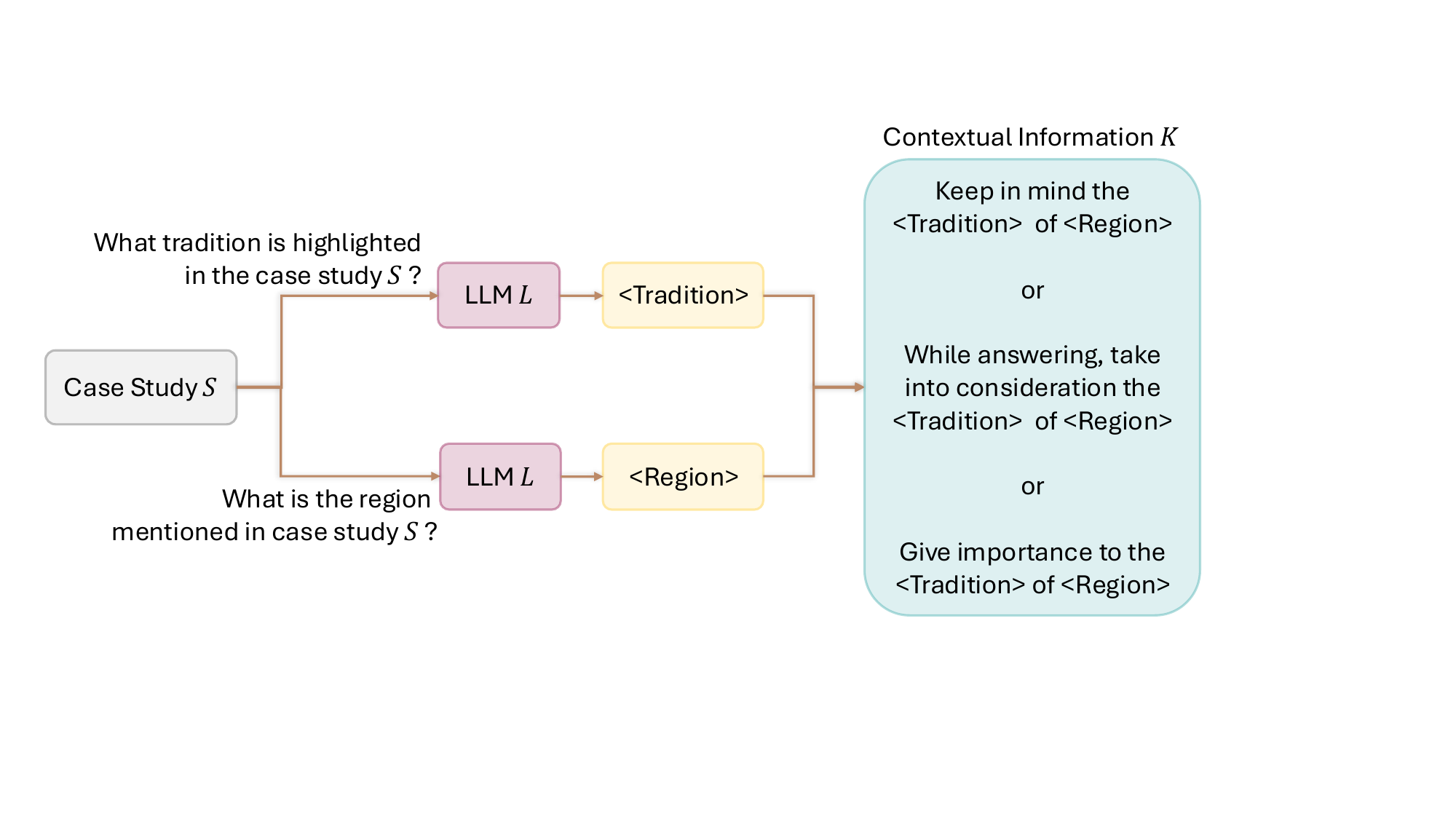}
\caption{Diagram depicts an automated approach for producing contextual information \MK. The Case Study \MS is provided as input to the LLM \ML, and is prompted to identify the tradition and region mentioned in \MS. The response from the LLM \ML is then combined with prefixes like ``Keep in mind'', ``While answering take into consideration'', or ``Give importance to'' to generate the final contextual information \MK.}
\vspace{-6mm}
\label{fig:automated_ce}
\end{figure}

\subsection{Automated Context Enrichment through LLM}
We experiment with a different setting where the LLM extracts information about the case study in automated fashion, such as the name of the tradition and the region to which the case study belongs. 
LLM is then prompted to focus on the extracted tradition and region to provide an appropriate response. The process of creating the contextual information \MK is presented in Figure \ref{fig:automated_ce}. 
% We prompt the LLM \ML to focus on a particular tradition while providing the response. 
Contextual information \MK provided to the LLM \ML for each case study is shown in Table \ref{tab:modified_prompts}. 
% \MK denote the knowledge base from the user, 
In this case, the prompt structure becomes (\MY , \ME) = \ML(\MK, \MS, \MI). 

% In this approach, we add the background  related to the case study.
% This will enrich the prompt with highly relevant knowledge which can be directly used to answer the prompt correctly. These background are extracted from \cite{cohn-chamar-kshatriya, avuncular-marriages-caste-ranking, mccormack-1958-sister, polyandry-kotas, kotas-indian, marriage-rules-north-india, bride-price-poona-village, lingayats-mcCormack, lingayats-renou, karve1990kinship, pollution-purity-harper, Gul_Shreya_2023, ambedkar2016cow, matrilineal-nayyar-nur-yalman}.
% Prompt $P_3$ is the prompt generated by function $F_3$. $F_3$ takes $P_1$ and adds context provided by the user (authors in this case). We call it the Knowledge base from user.

% \[
% \begin{aligned}
% P_3 &= F_3(P_1 + \text{KB(User)}) \\
% &\text{where KB(User) is the knowledge base provided by the User.}
% \end{aligned}
% \]

% In this approach we will implement the context enrichment in two levels. In the first level, that is, level - 1, we will introduce only the basic details about the case study. For example, if our case study is based on bride price that which place in a particular community in a particular region then we will add the details about the bride price and the community practicing it in the case study. For example we may introduce $(\alpha_1, \alpha_2)$ in the prompt and leaving out all the rest. This constitutes the level - 1 of the context enrichment by the user.

\begin{table*}[t]
\centering
\small
% \begin{tabularx}{\textwidth}{l|X}
\begin{tabular}{l|l}
\thickhline
\rowcolor{Gray} \textbf{No.} & \textbf{Added Prompt Instruction} \\
\thickhline
CS-1 & 
% From the above given scenario, What do you think which caste allows more decision making power and independence to their women?
Keep in mind the power dynamics of upper caste and the lower caste while answering the question. \\
% \hline
CS-2 & 
% Choose the correct option according to you from the following.
Keep in mind the avuncular marriages that take place in South India and the kinship structures in South India. \\
% \hline
CS-3 & 
% According to the above context, please choose the correct option.
% While answering, 
Think about the Kotas of Nilgiri Hills and also take into consideration their kinship structures and relations. \\
% \hline
CS-4 & 
% According to the above scenario, please tell me which side you think made the transaction and very precisely followed by a brief explanation of why a particular option was chosen (10-15 words).
While answering, take into consideration instances of bride price in lower castes. \\
% \hline
CS-5 & 
% From the above situation, tell me that it will be okay to marry the girl I am talking about keeping in mind all the cultural and traditional aspects and restrictions. Also 
Take into consideration the exogamy rules of a village in North India. \\
% \hline
CS-6 & 
% Please tell me if I was right or wrong in the above situation. Please choose one option from the following, followed by a brief explanation (10-15 words) as to why a particular answer was chosen.
While answering, consider the religious groups like Lingayats of South India. \\
% \hline
CS-7 & 
% Please tell me if I can marry that man, and according to you, will I face any opposition from the elders of the village?
Before answering, try to look into the village-to-village exogamy rules of Kishangarhi. \\
% \hline
CS-8 &
% I am unsure whether my actions were wrong and would appreciate your help in judging this matter. 
Make sure to accommodate the social importance of religion and religious practices in India. 
% Please choose the option you think is correct. 
\\
% \hline
CS-9 &
% Now I have to categorize the recorded instances into right and wrong acts according to popular beliefs of Hinduism. Also, 
View this case in light of variations present in the dietary habits of Brahmins across India. 
% Tell me in which category should I place the above discussed instance. You have 2 options to choose from. 
\\
% \hline
CS-10 & 
% According to you in the above scenario, choose the options below that you think might be the reason for her stay at her natal family.
While answering, try to accommodate the example matrilineal community of Nayyars. \\
CS-11 & 
Keep in mind the local traditions of Assam such as Bhekuli Biyah. \\
CS-12 & 
Give importance to the local rituals and traditions. \\
\thickhline
\end{tabular}
\vspace{1mm}
\caption{Additional information \MK provided for prompt enhancement.}
\label{tab:modified_prompts}
\end{table*}

\subsection{Results}

\subsubsection{Models Comparison} 

Table \ref{tab:all_results} presents the result for different prompting strategies. For \textit{vanilla} prompts, Gemini achieved the highest accuracy which is 83.3\%, but when we consider the accuracy of explanations, Gemini's performance dropped to 41.6\%. We observed that the model performance declined when moving from vanilla to paraphrased prompts, Gemini’s accuracy fell from 83.3\% to 50\%. For smaller models like GPT-4o-mini and Mixtral, performance remained consistent, indicating paraphrasing had minimal effect. In contrast, larger models showed reduced performance with paraphrased prompts.
When prompts are enriched via information extraction from LLMs, GPT-4o performed best with 83.3\% accuracy, followed by GPT-4o-mini at 58.3\%. Llama, Mixtral, and Gemini each achieved 50\% accuracy in this setting. The highest overall accuracy was observed when automated context enrichment is used to provide additional context. Under this strategy, Llama achieved 91.6\%, GPT-4o reached 83.3\%, Mixtral and Gemini both recorded 75\%, and GPT-4o-mini had 41.6\%. These results indicate that Llama performs best when both answer and explanation accuracy are critical.

In summary, the analysis shows that model accuracy is lowest when using paraphrased prompts, while the highest accuracy is achieved when automated method is utilised to enrich the prompts with additional context. This improvement is likely due to the fact that the LLMs are being guided to focus on a specific community and region while answering the case studies.
Our findings indicate that \textit{models with larger parameter sizes (approximately 70B), such as GPT-4o, Llama, and Gemini, benefit significantly from automated context enrichment}; in contrast, smaller models like GPT-4o-mini and Mixtral (around 7B parameters) do not show competitive performance even when additional context is given.

\subsubsection{Case Study Comparison} 
Context enrichment by the LLM resulted in incorrect answers for CS-5 and CS-7, both of which pertain to marriage rules within villages. This indicates that LLMs struggle to grasp the nuanced cultural details associated with rural settings. Even when the automated method of generating contextual information is used, most models still fail to generate correct explanations for these two case studies.
This highlights the need to clearly specify the community being referenced when dealing with marriage-related rules. The strongest performance is observed in the kinship and caste domains when automated context enrichment is applied, outperforming results obtained with vanilla prompts.
Overall, \textit{the models tend to perform reliably on caste and kinship scenarios but fall short in addressing the complexities of marriage customs, village norms, and religious contexts}.
% Context enrichment by the LLM gave all incorrect results for CS-5 and CS-7, both these case studies are related to the marriage rules in village. This shows that the LLMs fail to check for the intricacies specifically when it is related to village. 
% When the user explicitly, provides the context, still most of the models fail to provide correct explanation for CS-5 and CS-7.
% This demonstrates that for marriage rules, it is important to specify the particular community to whom the user is referring to. 
% The best performance is observed across the kinship and caste domain, when the context enrichment is done by the user as compared to using vanilla prompts. 
% Overall, \textit{the models perform well across Caste and Kinship, but fail to deal with the nuances of the marriage rules, village norms and religion.}

\subsection{Does using Indian Language for the Prompts Help?}

\begin{table*}[t]
\centering
\tiny
\begin{tabular}{l | c c c c c c c c c c c c | c c}
\thickhline
 \multirow{2}{*}{\textbf{Models}} & \textbf{CS-1} & \textbf{CS-2} & \textbf{CS-3} & \textbf{CS-4} & \textbf{CS-5} & \textbf{CS-6} & \textbf{CS-7} & \textbf{CS-8} & \textbf{CS-9} & \textbf{CS-10} & \textbf{CS-11} & \textbf{CS-12} & {\textbf{Accuracy}} & {\textbf{Accuracy}}\\
\cline{2-13}
& Hindi & Kannada & Tamil & Marathi & Hindi & Kannada & Hindi & Hindi & Bengali & Malayalam & Tripuri & Assamese &\textbf{\MY} &\textbf{\ME}\\
\thickhline
\rowcolor{Gray} \multicolumn{15}{c}{Vanilla Prompts} \\
\thickhline
gpt-4o & \gs \gc & \rs \rc  & \rs \rc  & \gs \gc  & \rs \rc   & \rs \rc  & \rs \rc  & \gs \gc  & \gs \gc & \gs \gc & \gs \rc  & \gs \rc & 58.3\% \blacksame & 41.6\% \blacksame \\ 
gpt-4o-mini & \rs \rc & \rs \rc & \rs \rc  & \rs \rc  & \rs \rc  & \rs \rc  & \rs \rc  & \rs \rc  & \gs \gc & \gs \gc & \rs \rc & \gs \rc & 25.0\% \reddown & 16.6\% \blacksame \\ 
llama-3.3-70b & \gs \gc  &\gs \gc & \gs \rc  & \gs \rc  & \rs \rc  & \rs \rc  & \rs \rc  & \rs \rc  & \gs \gc  & \gs \gc  & \rs \rc & \gs \rc & 58.3\% \blacksame & 33.3\% \greenup\\ 
mixtral-8x7b & \rs \rc  & \rs \rc  & \rs \rc & \gs \rc & \rs \rc  & \rs \rc  & \rs \rc  & \rs \rc  & \gs \gc  & \gs \gc  & \color{blue}{$\diamondsuit$} & \gs \rc & 33.3\% \reddown & 16.6\% \reddown\\ 
gemini-1.5-flash & \rs \rc  & \rs \rc  & \gs \rc & \rs \rc & \rs \rc & \rs \rc & \rs \rc & \gs \gc & \gs \gc & \gs \gc & \gs \rc & \gs \gc & 50.0\% \reddown & 33.3\% \reddown \\ 
\thickhline
\rowcolor{Gray} \multicolumn{15}{c}{Automated Context Enrichment through LLM} \\
\thickhline
gpt-4o & \gs \gc & \rs \rc  & \rs \rc  & \gs \rc & \rs \rc & \rs \rc  & \rs \rc & \gs \gc & \gs \gc  & \gs \gc & \gs \rc  & \gs \gc & 58.3\% \reddown & 41.6\% \reddown\\ 
gpt-4o-mini & \rs \rc & \rs \rc  & \rs \rc & \rs \rc  & \rs \rc  & \rs \rc & \rs \rc  & \rs \rc  & \gs \gc  &\gs \gc & \rs \rc  & \gs \gc  & 25.0\% \reddown & 25.0\% \reddown\\ 
llama-3.3-70b & \gs \gc  & \gs \gc  & \gs \gc  & \gs \gc  &\gs \gc  &\gs \gc  & \rs \rc  & \gs \gc  & \gs \gc  &\gs \gc  & \rs \rc & \gs \gc  & 83.3\% \reddown & 83.3\% \blacksame\\
mixtral-8x7b  & \rs \rc & \rs \rc & \gs \rc & \gs \gc & \rs \rc  & \gs \gc  & \rs \rc  & \rs \rc  & \gs \gc  &\gs \gc & \gs \gc & \gs \gc & 58.3\% \reddown & 50.0\% \reddown\\ 
gemini-1.5-flash &\rs \rc &\rs \rc  & \gs \gc & \gs \gc & \rs \rc & \rs \rc  & \rs \rc  & \gs \gc  & \gs \gc  & \gs \gc  & \gs \gc  & \gs \gc & 58.3\% \reddown & 58.3\% \reddown\\ 
\thickhline
\end{tabular}
\vspace{1mm}
\caption{Results from different LLMs when prompted in regional language. 
% \gs represents correct prediction, \gc represents correct explanation to the answer, \rs represents wrong prediction and \rc represents wrong explanation. The results are shown for two cases -- vanilla prompts and user context enriched prompts.
\greenup indicates the improvement when prompted in regional language as compared to English. \reddown indicates the worsening of the results after prompting in regional language. \blacksame represents same response, that is, no change. \diamondblue represents the cases where hallucinations were observed.}
\label{tab:regional}
\vspace{-5mm}
\end{table*}

% Prompts in the last section utilized English language, we hypothesize that prompting the LLM in the local regional language of the corresponding area might help in improving the result quality. 
So far, we conducted experiments using prompts in the English language.
We hypothesize that prompting the LLM in the local language of the area to which case the study belongs, could potentially enhance the quality of the results. For instance, ‘Kotas of Nilgiris’ case study pertains to Kollimalai Village, located in the state of Tamil Nadu. Consequently, the prompt was translated into state language Tamil. Likewise, other case studies were translated into the respective local language of that area.
% We now translate the given case studies in the local regional language of the area considered in the case studies. 
% in which the case studies were conducted. 
% We employed linguistic experts to translate the case studies. 
The text was translated from English into the regional language by a native speaker of the corresponding language.
% We utilized Google Translator\footnote{\url{https://translate.google.co.in/}} to translate the case studies from English to regional language. \textcolor{blue}{Authors and the sociologists are familiar with local languages, and validated the text before feeding into LLMs.}
For example, the case study of ``The Unfamous Dowry: Bride Price'', belonged to a village in Maharashtra. Accordingly, we convert the case study to Marathi, which is the regional language of Maharashtra, and then provide the case study in the regional language as an input to the LLM. 
Table \ref{tab:regional} mentions the local regional language to which the case study was translated and the results obtained after the translation.
We experimented with local languages for two reasons: 
(i) More people now use LLMs in native languages, with commercial models adding Indian language support. It is crucial to evaluate whether these models can comprehend cultural nuances in local language; and 
(ii) During pre-training, LLMs may have encountered local language content reflecting the regional customs. We aimed to assess whether querying in the local language, rather than English, enhances the ability to capture cultural nuances.

From Table \ref{tab:regional}, we observe that in the majority of instances, the model's performance deteriorated when prompted in a regional language as compared to when prompted in English. 
% This suggests that 
The use of regional languages in prompts adversely affects the model’s ability to generate accurate or contextually appropriate responses.
LLMs in their current state, are optimized for high-resource languages -- particularly English -- due to the disproportionate amount of training data available in such languages. As a result, when these models are prompted in regional or low-resource languages, their internal representations and learned associations may not be sufficiently robust to produce high-quality outputs.
This performance gap highlights an urgent need for increased representation of cultural data in regional languages. Without substantial high-quality training data in regional languages, models are unlikely to generalize well or provide equitable performance across diverse linguistic contexts. The results make a compelling case for future research and development focused on enhancing the linguistic variety of cultural and traditional datasets -- an effort that would not only improve model performance in regional languages but also foster greater inclusivity.

\section{Related Work}
%Prediction systems tend to work in conjunction with the organisation and are likely to reinforce the organization's existing biases and behaviors rather than correcting or changing them \cite{barocas2023fairness}. 
%By its inherent nature, human decision making have a lot of ``noise'' (variations) \cite{noise_human_decision}, whereas Machine Learning (ML) models tend to homegenize the decision making \cite{Boyd2012-qb}.
%ML models when deployed in decision making tend to ignore the noise and generalise the decision making, which in turn leads to erasure of the less prominent aspect \cite{barocas2023fairness}.
%Predictive systems inherit the structural discrimination from the organisation of which they are a part of \cite{barocas2023fairness, racial-bias-health}, such as targeted advertisements reinforce stereotypes \cite{Merrill_undated-al}.

Prediction systems often operate in tandem with organizational structures, making them more likely to amplify existing biases and behaviors rather than challenge or correct them. Machine Learning (ML) models deployed in decision-making processes tend to generalize outcomes by overlooking nuanced or less prominent aspects, leading to the erasure of minority perspectives \cite{barocas2023fairness}. Moreover, predictive systems inherit the structural discrimination embedded within the organizations they serve \cite{racial-bias-health}. For example, targeted advertising algorithms frequently perpetuate stereotypes, further entrenching societal biases rather than mitigating them~\cite{Merrill_undated-al}.

% Due to these phenomena, the bias and segregation are also reflected in social media as well.\cite{Boyd2012-qb}
LLMs are also a variant of predictive systems and treats the observable phenomena as numbers which might not capture the real meaning of cultural aspect \cite{barocas2023fairness}. 
Recent studies have highlighted that LLMs struggle to grasp cultural nuances, often displaying an english-centric bias and limited proficiency in regional languages \cite{dawson-evaluating-cultural-awareness-llms, blodgett-etal-2020-language, xu2024exploringmultilingualconceptshuman}. 
While LLMs can define culture, they perform poorly in reasoning, possibly due to memorizing cultural information rather than truly understanding its complexities \cite{liu-multilingual-llms-culturally-diverse-reasoners}. 
Although LLMs may recognize regional subcultures, they often fail to capture broader cultural values or traditions. 
They lack the comprehension of localized cultural intricacies \cite{llms-represent-values-cultures}, and are prone to misrepresenting and misinterpreting cultural contexts \cite{prabhakaran2022culturalincongruenciesartificialintelligence}.
A framework is proposed to enhance the understanding of cultural differences in LLMs \cite{li2024culturellmincorporatingculturaldifferences}. 
The concept of Representation Engineering (RepE) demonstrates that abstract concepts within LLMs can be extracted as vectors, which can be leveraged to improve the models cultural understanding \cite{zou2023representationengineeringtopdownapproach}. 
LLMs favor western cultural values, leading to significant inequity, and addressing this requires embracing cultural diversity \cite{Schooler1983-hl, Cao2023-yi, Johnson2022-vu, Masoud2023-tc, liu-multilingual-llms-culturally-diverse-reasoners, Mohamed2020}. 
These biases can potentially be mitigated through techniques such as prompt engineering and pre-training, both of which have been shown to deliver promising results in some cases \cite{kovač2023largelanguagemodelssuperpositions, wang2024countriescelebratethanksgivingcultural}. 

% Text-to-image models have been observed to produce results that often reflect broad generalizations of specific queries -- when tasked with generating a descriptive image of a market in Varanasi, India, the output produced was a depiction of a generic Indian market rather than one accurately reflecting the unique characteristics of Varanasi. 
% This highlights the inherent tendency of generative models to favor dominant or generalized perspectives \cite{Qadri_2023}. A significant challenge lies in the models inability to reconcile Western cultural frameworks with the intricate and distinct cultural values of the East. This cultural incongruence often leads to a failure in capturing the nuanced and contextual aspects of non-Western societies \cite{prabhakaran2022culturalincongruenciesartificialintelligence, sambasivan2021reimagining, sambasivan2020nonportability}. There is an immediate need to re-contextualize the data and model evaluation with amplified attention towards the under-represented aspects of the culture \cite{sambasivan2020nonportability}.
% These generative models also perpetuate the existing caste dynamics as well \cite{Qadri_2023}.

Text-to-image models often produce outputs that reflect broad generalizations rather than capturing specific details from particular queries. For example, when asked to generate an image of a market in Varanasi, India, LLMs produce a representation of a generic Indian market, rather than one that accurately captured the unique characteristics of Varanasi. This demonstrates a tendency of generative models to prioritize dominant or generalized viewpoints \cite{Qadri_2023}. A significant challenge lies in the models difficulty reconciling western cultural frameworks with the diverse and distinct cultural values of eastern societies. This cultural mismatch often results in a failure to capture the nuanced and contextual aspects of non-western cultures \cite{prabhakaran2022culturalincongruenciesartificialintelligence, sambasivan2021reimagining, sambasivan2020nonportability}. Therefore, there is a need to re-contextualize data and model evaluations, with increased focus on the under-represented cultural elements \cite{sambasivan2020nonportability}. Additionally, these generative models can reinforce existing caste dynamics \cite{Qadri_2023}. 
% A concerning revelation is that
% LLMs tend to replicate societal issues, where dominant cultures overshadow and diminish local traditions \cite{niszczota2024largelanguagemodelsreplicate}. In this work we explore whether LLMs possess knowledge about India's sub-cultures and lesser-known traditions, and examines their ability to provide appropriate reasoning. 
LLMs often reflect societal issues, where dominant cultures overshadow and marginalize local traditions \cite{niszczota2024largelanguagemodelsreplicate}. 

% These biases in the LLMs are further propagated as there is lack of diversity and inclusion in AI practices. These kind of biases emerge due to numerous drawbacks including paucity of availability of datasets with diverse backgrounds and which are truly representing the society.\cite{diversity-inclusion-in-AI}

The existing literature examines biases in  text generation \cite{10.5555/3716662.3716702}, image generation \cite{10.5555/3716662.3716703}, and other AI tools \cite{10.1145/3613904.3642669}, often comparing disparities between the Global North and South \cite{bhatt-diaz-2024-extrinsic, ignat2024, rao2025normad}. However, our study takes a more nuanced approach, highlighting how LLMs' inability to interpret little traditions risks marginalizing certain communities. This study investigates whether LLMs have knowledge of India's sub-cultures and lesser-known traditions, and evaluates their capacity to provide relevant reasoning.
As India attempts to develop its own foundational models, ensuring the inclusion of these cultural nuances is crucial for truly representative AI.

\section{Conclusion}
In this work, we explored the ability of LLMs to comprehend the little traditions of India. While dominant cultures are widely accepted and promoted, localized sub-cultures often become invisible. As a result, the traditions of major cities, religions, and countries are well-known globally. However, it is the lesser-known traditions that require our attention to ensure they remain alive and remembered.
Our study focuses on states with distinct socio-cultural practices. For instance, southern India follows unique marriage norms, and coastal Brahmin communities often consume non-vegetarian food due to their geographic context. Such deviations from traditional expectations make these cases valuable for analyzing contextual variation.

% In case of translating the prompts in regional languages and then passing to the models, there is no significant difference in the accuracy. In case of GPT-4o-mini the accuracy decreased. Low parameter models perform worse when the translated prompts are passed. 

% LLMs when asked to answer a question based on a situation, usually pick the response that supports the cultural aspect related to the dominant culture.
% The accuracy worsened when the scenarios are presented after paraphrasing (Table \ref{tab:all_results}). Improvements were observed when context enrichment approach was implemented. In case of context enrichment from LLM, though the accuracy improved, 25/50 instances were correctly answered, but it is still not satisfactory. The major improvement was observed when some hint in the form of context is given is given by the user, i.e., 32/50 instances were correctly answered. 
% The models chose the options that were accommodating the cultural aspects related to the little traditions. 
% The results carry several implications. 

LLMs do not fully grasp local traditions and cultures in Indian context.
% The highest explanation accuracy achieved is  41.6\%, when the \textit{vanilla} prompts are passed, which is alarmingly low. 
Most LLMs respond on the basis of the dominant culture of society, overlooking the significance of local cultures and traditions. 
Extra context needs to be provided to get better results. Often, models hold knowledge about the culture and traditions but are not able to reason with it when asked to do so (as can be seen in \textit{vanilla} setup).
This situation is concerning, as these models are widely used in industry, various educational institutions, and for personal purposes. Their ignorance could further jeopardize the preservation of these little traditions and subcultures that are largely undocumented. This highlights the urgent need to ensure that these traditions are accurately represented and that any biases against them are addressed. Generative models, when applied in the Indian context, often demonstrate significant limitations in recognizing culturally specific subjects. In numerous instances, these models exhibited a complete inability to comprehend such cultural nuances \cite{Qadri_2023}.

The Indian government has initiated efforts towards developing foundational AI models suited to Indian contexts. 
Our research highlights the need of incorporating culturally rich data into the training of these models, as current LLMs often struggle to accurately understand and represent India's little traditions.
Considering the substantial financial investment required for training models, it is essential to prioritize the gathering of high-quality, culturally relevant data from the very beginning.
% Given the high cost of model training, collecting high-quality cultural data from the outset is crucial. 
Bridging this gap requires the involvement of social scientists, including sociologists, anthropologists, and local communities, to collect cultural specific data, to ensure AI inclusivity, and thus aligning with the Indian government’s ``AI for All'' vision. 
Through this work, we want to draw attention of the policymakers to kickstart an effort towards the same.

\noindent \textbf{Generalizability of findings:}
India is a culturally rich and diverse nation, where traditions, customs, and social practices vary significantly across different regions.
Given this immense cultural diversity, compiling a comprehensive list capturing the full spectrum of culture is impractical because the traditions are often hyper-local and orally transmitted through generations.
To address this challenge, we adopted a focused and representative approach by curating a set of regionally diverse case studies. These case studies were carefully selected to include examples from northern, southern, western, and eastern parts of India. 
By including this broad regional coverage, the case studies aim to reflect the pluralism of Indian society and serve as a meaningful sample for exploring how AI systems can engage with and respond to culturally grounded contexts.

\bibliographystyle{IEEEtran.bst}
\bibliography{ref}

% \appendix
% \input{appendix}

% \newpage
\begin{IEEEbiography}
% left bottom right top
[{\includegraphics[trim={0 200 0 100},width=1in,height=1.5in,clip,keepaspectratio]{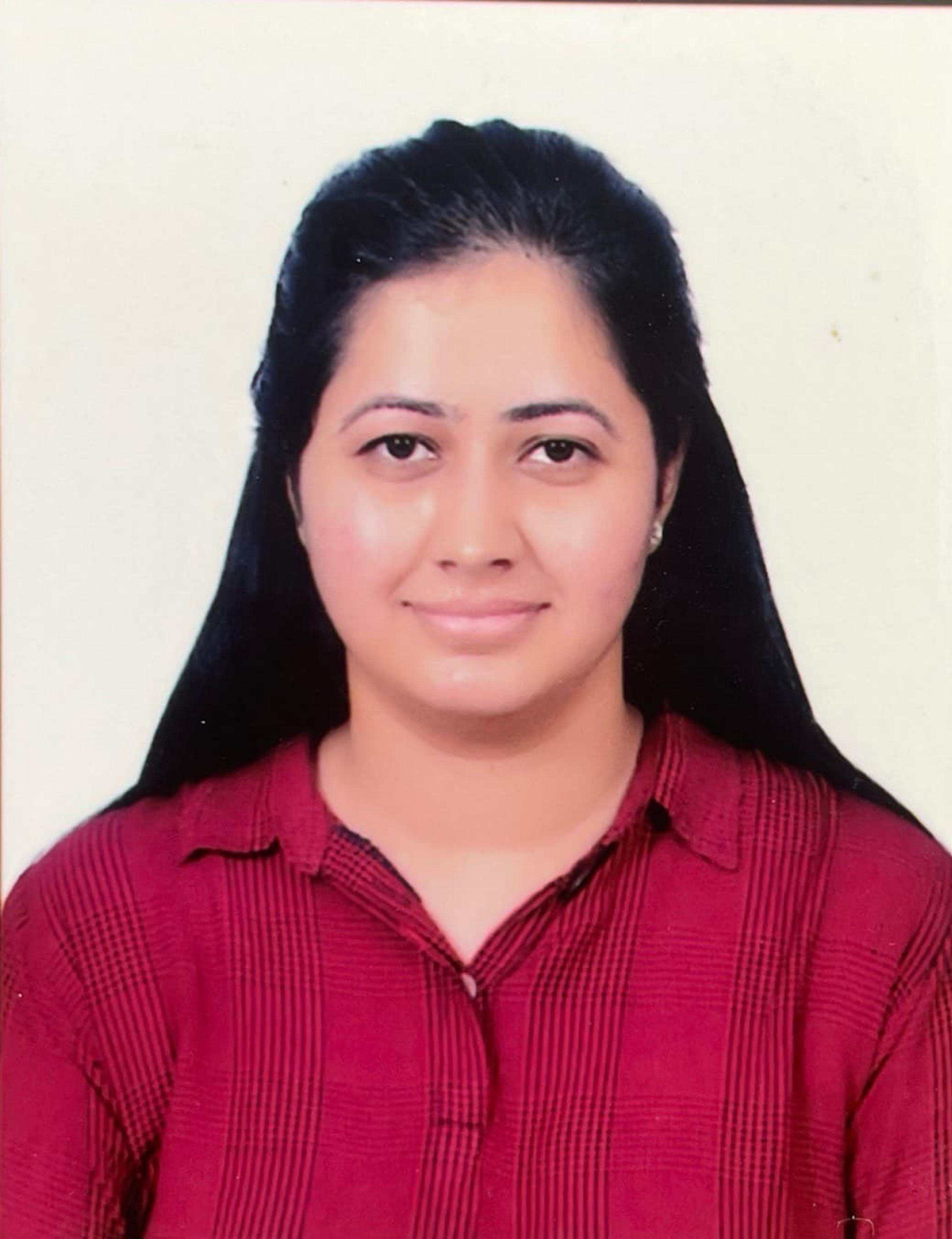}}]
{Garima Chhikara}
is as Assistant Professor in Delhi Technological University in Computer Science and Engineering Department. She is also a PhD student in School of IT at IIT Delhi, advised by Prof. Abhijnan Chakraborty. Her research interests include Fairness in Summarization, Social Network analysis, Computational Social Science. She obtained Masters from IIIT Delhi under the supervision of Prof. Vinayak Naik.\end{IEEEbiography}

\begin{IEEEbiography}
[{\includegraphics[width=1in,height=1.5in,clip,keepaspectratio]{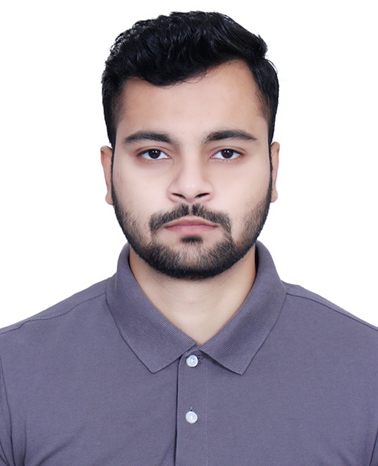}}]
{Abhishek Kumar} is a Technical Lead at HCLTech, working in the domain of Generative AI. He earned his master's degree from Delhi Technological University. 
\end{IEEEbiography}

\begin{IEEEbiography}
[{\includegraphics[width=1in,height=1.5in,clip,keepaspectratio]{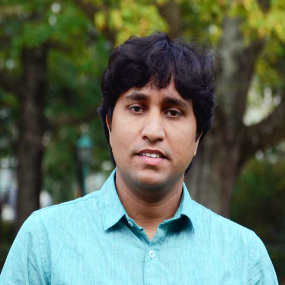}}]
{Abhijnan Chakraborty} is an Assistant Professor in the Department of Computer Science and Engineering at Indian Institute of Technology (IIT) Kharagpur. Earlier, he was an Assistant Professor at IIT Delhi in the Department of Computer Science and was also associated with the School of Artificial Intelligence. In 2022, he was inducted as a Young Associate of the Indian National Academy of Engineering (INAE). Before IIT Delhi, he worked for 2.5 years at the Max Planck Institute for Software Systems (MPI-SWS) as a post-doctoral researcher. He obtained his PhD from IIT Kharagpur, where he was jointly advised by Prof. Niloy Ganguly from IIT Kharagpur and Prof. Krishna Gummadi from MPI-SWS. During PhD, he was awarded the prestigious Prime Minister's Fellowship for Doctoral Research and Google India PhD Fellowship.
\end{IEEEbiography}

\end{document}